# Fluid Grey 2: How Well Does Generative Adversarial Network Learn Deeper Topology Structure in Architecture That Matches Images?

Yayan Qiu[1, *], Sean Hanna[1]

[1] Bartlett School of Architecture, University College London, London WC1E 6BT, UK; yayan.qiu.21@alumni.ucl.ac.uk; s.hanna@ucl.ac.uk

[*] Correspondence: yayan.qiu.21@alumni.ucl.ac.uk

## Author Note

## Abstract

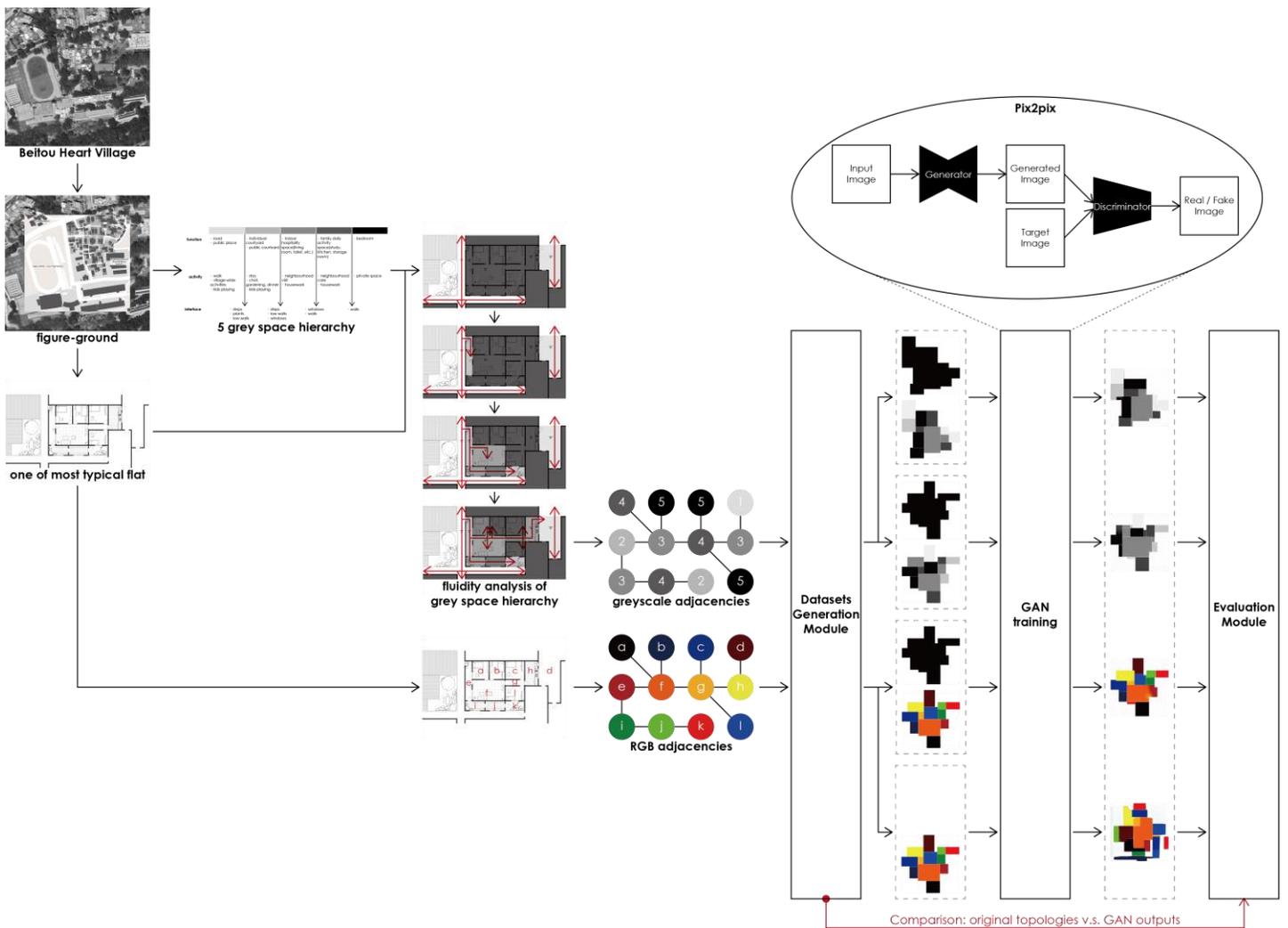

Taking into account the regional characteristics of 'intrinsic' and 'extrinsic' properties of space is an essential issue in architectural design and urban renewal, which is often achieved step by step using image and graph-based GANs. However, each model nesting and data conversion may cause information loss, and it is necessary to streamline the tools to facilitate architects and users to participate in the design. Therefore, this study hopes to prove that I2I GAN also has the potential to recognize topological relationships autonomously. Therefore, this research proposes a method for quickly detecting the ability of pix2pix to learn topological relationships, which is achieved by adding two Grasshopper-based detection modules before and after GAN. At the same time, quantitative data is provided and its learning process is visualized, and changes in different input modes such as greyscale and RGB affect its learning efficiency. There are two innovations in this paper: 1) It proves that pix2pix can automatically learn spatial topological relationships and apply them to architectural design. 2) It fills the gap in detecting the performance of Image-based Generation GAN from a topological perspective. Moreover, the detection method proposed in this study takes a short time and is simple to operate. The two detection modules can be widely used for customizing image datasets with the same topological structure and for batch detection of topological relationships of images. In the future, this paper may provide a theoretical foundation and data support for the application of architectural design and urban renewal that use GAN to preserve spatial topological characteristics.

## Highlights

Prove pix2pix can auto-learn spatial topological relationships
General method for testing topological learning ability of Image-to-image Translation
Visual analysis of the topological learning process of Image-to-image Translation

## Abbreviations

| | |
|---|---|
| AI | Artificial Intelligence |
| ANN | Artificial Neural Network |
| cGAN | Conditional Generative Adversarial Network |
| CNN | Convolutional Neural Network |
| DL | Deep Learning |
| GAN | Generative Adversarial Networks |
| GCN | Graph Convolutional Network |
| GNN | Graph Neural Network |
| I2I | Image-to-image Translation |
| ML | Machine Learning |

**Fluid Grey 2: How Well Does Generative Adversarial Network Learn Deeper Topology Structure in Architecture**

**That Matches Images?**

## 1. Introduction

### 1.1. Regional Characteristics of Space

Preserving the regional characteristics in architectural design and urban space has always been a major issue in the field of architecture. It requires balancing the 'intrinsic' and 'extrinsic' properties of space—the former is visible and mainly presented through geometric properties, such as shape, size, volume, pattern and texture; the latter consists of invisible structural relationships, mainly the arrangement and topology of elements [43] . However, most modern theories only focus on one of them and find it difficult to balance the two [43] . On the one hand, critical regionalism, which historically emphasizes regional characteristics, places too much emphasis on the intrinsic properties of space. In 1981, the term critical regionalism was first proposed by Alexander Tzonis and Liane Lefaivre; it is a design and planning method that can promote the ecological, social and intellectual singularity and diversity of regions [42] . Subsequently, in 1983, Kenneth Frampton further proposed that its basic strategy is to mediate the influence of universal civilization with elements that are indirectly derived from specific local characteristics, such as culture, place–form, natural environment, vision, etc. [14]. However, it is difficult to make general statements about future impacts when only considering the 'intrinsic' properties of space [43] . On the other hand, urban planning, from ancient practice to modern urban network theory, has always been overly biased towards the extrinsic properties of space. The earliest practice of urbanism can be traced back to Pope Sixtus V's reform of the ancient Roman city in the 16th century, which strengthened the symbol of religious power by strengthening the urban axis with obelisks and reorganizing streets [2]. From 1852 to 1870, Georges-Eugène Haussmann carried out an urban transformation plan for Paris based on this, thus forming the radial Paris road network of the Champs-Élysées [2]. It was not until the 20th century that urban network theory gradually developed a theoretical explanation of the urban linear topological hierarchy as industrialization in modernism, represented by the mass production of Ford cars, which caused a general change in transportation methods. In 1927, Ludwig Hilberseimer first proposed the street hierarchy, which embeds the importance of each road type into the network topology [16] . Jane Jacobs criticized that, although it helped promote the formation of a clear road network in cities, it also pushed cities towards being car-oriented and caused the demotion of pedestrians [24]. In the 1970s, Bill Hillier and Julienne Hanson et al. developed space syntax [17] [18] [19] [20]  based on network theories, which is a mathematical street network model for calculating topological spatial relationships [43] . In the 21st century, Kim Dovey and Elek Pafka developed the urban DMA [12]  inspired by assemblage theory and Jane Jacobs [24] , arguing that cities rely on the complex synergy of density, mix, and access. The mix element of this explores the distribution of the three functions of visiting, life, and work in cities, proving that land use is multiple in any good city [12] . However, it only focuses on these three functions and lacks an in-depth understanding of functions and topology as well as a horizontal comparison of regional characteristics. Space syntax and urban DMA are the few theories that use systematic methods to quantify and compare the similarities and differences in different urban characteristics. They horizontally analyse and separately compare how cities in different regions achieve common points in accessibility and mixed uses in quantitative data under different modes from two perspectives: the richness of the road network system and mixed land use zoning. The success of these two theories also verifies that combining the computing power advantages of new technology will be more conducive to achieving a balance between the 'intrinsic' and 'extrinsic' properties of space in buildings.

### 1.2. Generative Adversarial Network (GAN) in Architecture

With the rise of machine learning, people have also tried to apply it to the field of architecture to make up for the limitations of humans compared to computers in terms of computing power, objectivity, etc. In the large field of Artificial Intelligence (AI), AI is any technique that enables computers to mimic human behaviour. It encompasses Machine Learning (ML), which is the study of computer algorithms that improve automatically through experience and have the ability to learn without explicitly being programmed[31] . Deep learning (DL) allows computational models that are composed of multiple processing layers to learn representations of data with multiple levels of abstraction [26] [1] , which is a subset of machine learning [3] . Generative adversarial network (GAN) is an important derivative branch of deep learning and has been the core research direction in the field of machine learning since 2015. GAN is derived from DL, its main difference is that the discriminator is used to replace the calculation of DL refinement reconstruction error, so it can be regarded as an improvement on the encoder [7] . GAN is an analogous type of idea generated to model animal behaviour by researchers around 2013[4] . It was formally proposed by Goodfellow and his colleagues in 2014 based on the inspiration of zero-sum game, and the basic idea behind it is to learn the probability distribution of training samples.[15] . GAN mainly allows two networks (generator network G and discriminator network D) to compete with each other, and captures the probability distribution of real samples in the dataset in G, and then adds random noise to transform it into fakes[7] . The basic idea of GAN network deployment is aimed at unsupervised machine learning technology, but it has also been proven to be a better solution for semi-supervised and reinforcement learning[1] . Conditional Generative Adversarial Network (cGAN) is a deep learning method that applies conditional settings and is an extension of GAN[30] . Its core is that both the generator and the discriminator are conditioned on some auxiliary information, that is, the conditional settings are integrated into the generator and the discriminator, which is also the main difference between it and GAN[30] . The application branches of Generative Adversarial Networks are mainly: Classification and Regression, Image Synthesis, Image-to-image translation (I2I), Super-resolution[8] . Among them, the application of Image-to-image Translation (I2I) is very rich and mature, whose goal is to convert the input image from the source domain to the target domain while retaining the internal source content and transferring the external target style[34] . Since cGAN is very suitable for converting input images to output images, which is a recurring theme in computer graphics, image processing and computer vision[8], therefore, pix2pix, as the original origin of the field of I2I, uses cGAN to solve the problem of I2I.

The application of GAN in architecture is mainly divided into two categories: Image-based Generation and Graph-based Generation[35] . The data types they learn correspond to the 'intrinsic' and 'extrinsic' properties of space respectively. Image-based Generation mainly uses pix2pix[23] created by Phillip Isola et al. in 2017 based on the cGAN[30] to solve the I2I problem, which is also the most widely used type in the field of architecture. The difference in its framework is that nothing is application-specific, which makes their setup tremendously simpler than others. Specifically, pix2pix achieves this versatility by choosing different architectures for the generator and the discriminator. They use the "U-Net"-based architecture[38] as the generator and the convolutional "PatchGAN" classifier[27] in the discriminator, which only penalizes the structure at the scale of image patches[23]. Although in 2018, Huang Weixin and Zheng Hao found that pix2pix had a certain degree of fault tolerance when applying it to floor plan drawing, and thus speculated that machine learning algorithms can mine abstract concepts from concrete entities and extract accurate standards from fuzzy understandings like humans [22] . However, they simply let pix2pix fill in the furniture design based on the functional floor plan, which only adds details to the picture without involving topological cognition. In 2020, Chaillou, S. used ArchiGAN to expand the scalability of pix2pix based on the work of predecessors, superimposing three models throughout the three steps of architectural floor plan design: building footprint massing; program repartition; furniture layout [5] . These applications actually use pix2pix as a filling tool, mainly for 2D recognition and filling of the plan or facade of the building — specifying the filling area in the source dataset and the filling target in the target dataset. Although pix2pix can achieve a high degree of realism in image learning, it still has limitations in topology, 3D and output diversity, and its algorithm also has generalization limitations [37] .

On the other hand, graph-based GAN is more directly concerned with learning and preserving topological structures, such as convolutional neural network (CNN) or graph neural networks (GNN) [39] or graph convolutional network (GCN) [25] in artificial neural network (ANN). GNN is a class of artificial neural networks for processing data that can be represented as graphs [39] . GCN is a method for semi-supervised learning of graph-structured data, which is an effective variant of CNNs that operate directly on graphs [25] . In 2019, Manandhar, D. et al. first used GNN [39] to learn the structural similarity of UI layouts with rectangular boundaries [29] . In the following years, many other attempts to learn architectural graphs have emerged, such as RPLAN [45] and Graph2Plan [21], etc. [32] [33] [11] [35] [41] , which can be divided into optimization-based and learning-based approaches [5]. The most well-known and applicable one is the House-GAN series [32] [33] , which uses a relational generator and discriminator, where the constraint is encoded into the graph structure of their relational neural networks. It is a novel graph-constrained house layout generator, a new generative model that can automatically generate house floor plan configurations directly from bubble graphs. More specifically, the Conv-MPN [47] used by HouseGAN is a variant of GNN, which learns to infer the relationship between nodes by exchanging messages. Since then, graph-based research has also been limited to small scales and 2D. There are also a few studies that extend it to the spatial domain, such as 3D GCN [9] and Building-GAN [6] , but their building layout programs are idealized and do not consider the influence of the external environment [6] .

Therefore, it can be found that many applications of GAN are still based on the technical perspective of computer science, so their application in architecture is not so suitable. Therefore, in order to respond to the need to preserve regional characteristics in the field of architecture, it is necessary to focus on GAN that combines image and graph. The current way to achieve this goal is to nest image and graph-based models. However, each nesting or conversion may cause information loss, so the merging and simplification of methods is a more ideal situation. Therefore, based on the existing models and gaps, there are two solutions: one is to discover the possibility of graph learning in image-based generation; the other is to optimize graph-based generation and add more details. Since most of the data and information obtained in the field of architecture are based on images, and image-based methods are more convenient and widely spread than graph-based methods, this not only makes image-based GAN still have a wider architects users than the latter, but also makes it easier to further use it as a convenient tool for co-design to achieve public participation. Therefore, it is necessary to supplement image-based generation with a test method that retains regional spatial characteristics in the field of architecture.

In summary, Taking into account the regional characteristics of 'intrinsic' and 'extrinsic' properties of space is an essential issue in architectural design and urban renewal, which is often achieved step by step using image and graph-based GANs. However, each model nesting and data conversion may cause information loss, and it is necessary to streamline the tools to facilitate architects and users to participate in the design. Therefore, this study hopes to prove that I2I GAN also has the potential to recognize topological relationships autonomously. Specifically, it is divided into three sub-questions:

1) Can GANs, especially pix2pix, autonomously learn the topological structure in images without topological cues?

2) How effective is it in learning topological structures?

3) What is the impact of greyscale and RGB modes on its learning process?

Therefore, this research proposes a method for quickly detecting the ability of pix2pix to learn topological relationships, which is achieved by adding two Grasshopper-based detection modules before and after GAN. At the same time, quantitative data is provided and its learning process is visualized, and changes in different input modes such as greyscale and RGB affect its learning efficiency. There are two innovations in this paper: 1) It proves that pix2pix can automatically learn spatial topological relationships and apply them to architectural design. 2) It fills the gap in detecting the performance of Image-based Generation GAN from a topological perspective. Moreover, the detection method proposed in this study takes a short time and is simple to operate. The two detection modules can be widely used for customizing image datasets with the same topological structure and for batch detection of topological relationships of images. In the future, this paper may provide a theoretical foundation and data support for the application of architectural design and urban renewal that use GAN to preserve spatial topological characteristics.

## 2. Methods

This paper focuses on the analysis and evaluation of the extent to which pix2pix can learn the topological structure of building planes. By adding pre- and post-evaluation modules before and after pix2pix training, and combining the output images, loss function, and adjacencies learning rate obtained by statistical extraction, this study further analyses how the turning point of the training curve will correspond to the topological changes in the output image. The process is divided into four steps (Figure 1):

**1) Generate Datasets:** Use C# and Magnetizing[13] plug-ins in Grasshopper to write a dataset generation module, generate floor plans with the same topological relationship as the pix2pix dataset based on the sample floor plans, and record their original topological structures.

**2) GAN Training:** Use Pix2pix to train two datasets in greyscale and RGB formats respectively.

**3) Evaluation:** On the one hand, analyse the loss curve of pix2pix. On the other hand, the images generated by pix2pix are imported into the topology evaluation module written in Grasshopper, and combined with the original topology structure recorded in the first step, the number of spatial neighbours and the learning rate of the pix2pix output results are analysed and evaluated.

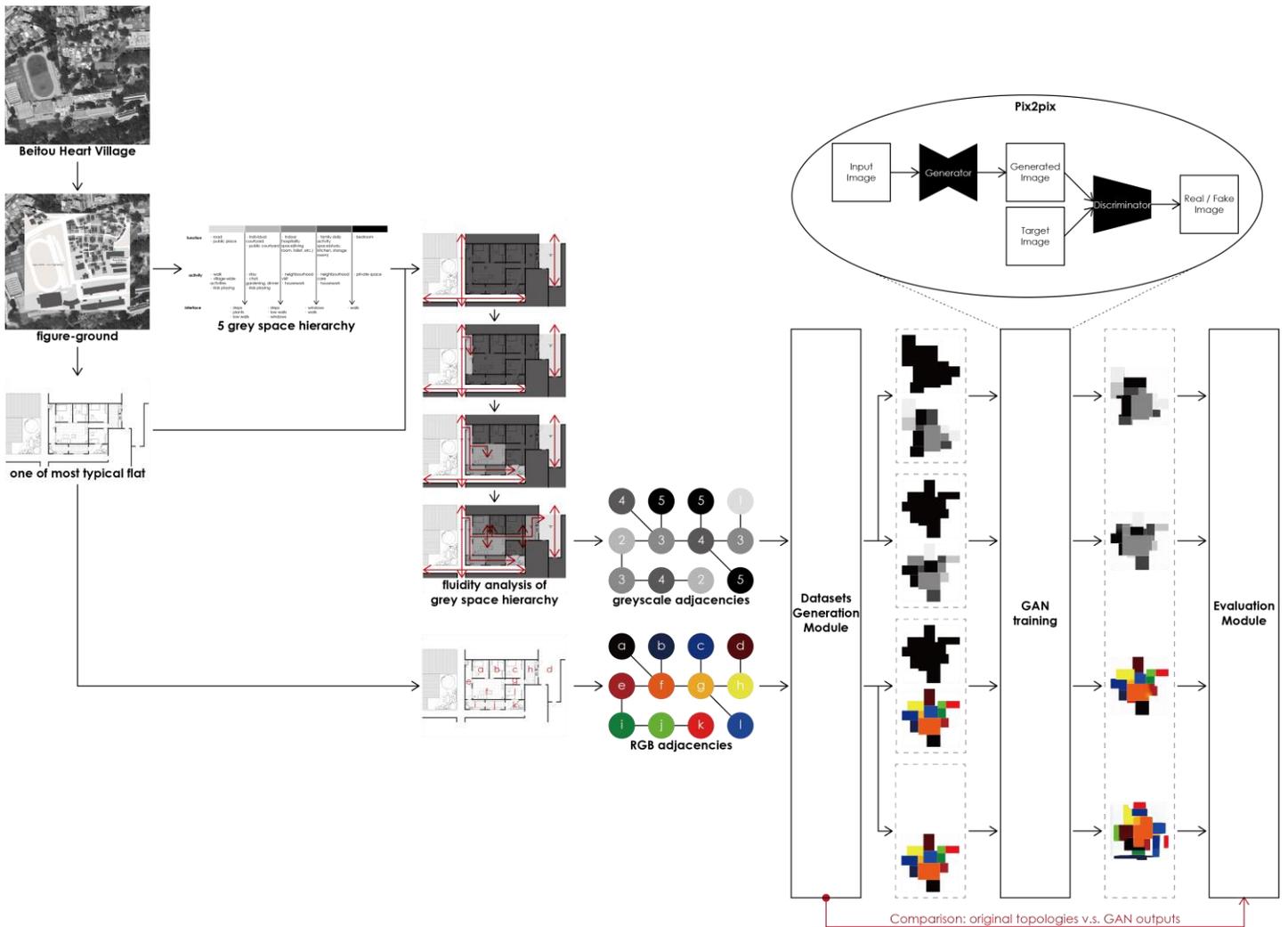

**Figure 1. Research Progress Diagram**

## 2.1. Generate Datasets

The author found that the reason why Beitou Heart Village, located in the northern part of Taiwan Province, China, is famous for its harmonious neighbourhood relations is that it has very rich spatial hierarchies (Figure 2) [36] . Therefore, the starting point for selecting the dataset in this paper is to retain the regional topological characteristics of Beitou Heart Village through pix2pix. This study uses the most typical family in the village as a sample and extracts the spatial topological relationship from it as the source of the dataset. Among them, the spatial topological relationship of function is used as the colour experimental group. The level division of Grey Space is the depth of spatial topology and the influence of actual usage, environment and human factors. The grey space hierarchy of Beitou Heart Village can be summarized into five types: 1) roads and public squares; 2) personal or small-scale public courtyards; 3) indoor reception space; 4) family daily life space; and 5) bedrooms (Figure 3). The higher the grey level, the stronger the privacy. We can analyse how this family changes the degree of openness of the space through the switches on the interface, achieving a continuous fluid of grey space and supporting diverse neighbourhood mutual assistance (Figure 3) [36] .

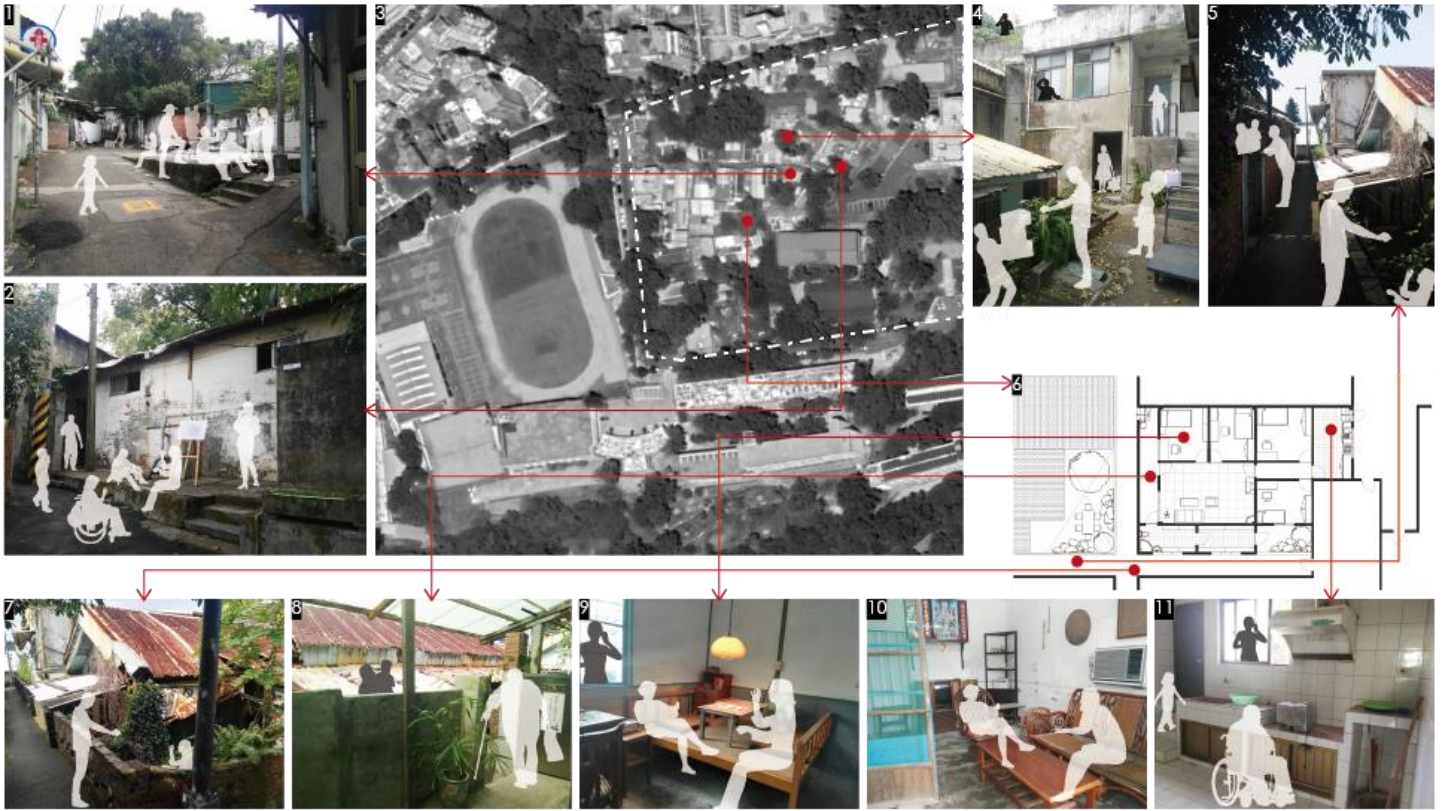

**Figure 2. Photos and Usage Analysis Collage of Beitou Heart Village [36]**

*Note.* The collage illustrates Beitou Heart Village's main modes of neighbourly communication in its public and indoor spaces. Subfigures (1–5) are the main public spaces of the village. Subfigure (1) is the largest public square in the village, and Subfigure (2) is the public hot spring bath. The height difference between them can provide multi-level leisure space. Subfigures (5–11) show the environment around the case house selected for this study. All on-site photos were taken from February to March 2019 by Yayan Qiu, the author of this paper. The satellite map in Subfigure (3) is from Google Earth.

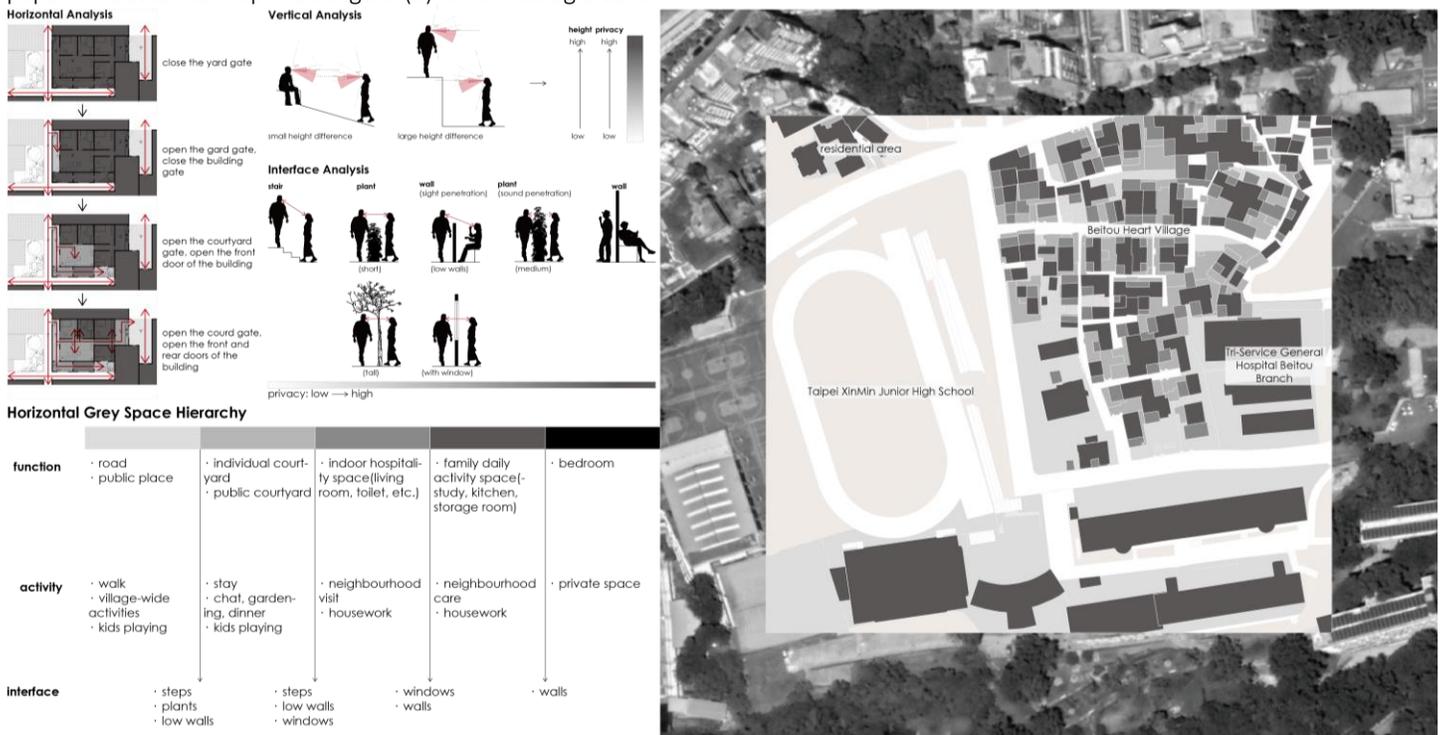

**Figure 3. Analysis of Grey Space Hierarchy(left) and Overall Figure-ground of Beitou Heart Village(right) [36]**

*Note.* The left side analyses the horizontal and vertical grey space levels, interfaces and flow patterns of Beitou Heart Village, respectively. On the right is the overall figure-ground of the entire village based on the five grey space levels obtained from the analysis. The satellite

map used is from Google Earth, and the topographic map numerical image file is from the Department of Urban Development, Taipei City Government [10] .

In Grasshopper, use C# and plug-ins such as Magnetizing[13] to write a dataset generation module. The datasets are generated based on the spatial adjacency relationship of the sample family floor plan, which is also recorded as the original topological structure. The dataset of pix2pix is divided into two groups: Input Image and Target Image (also known as Ground Truth)—the former is the constraint given to it, as the training source; the latter is the appearance that the training result is expected to be close to, as the training target. Since the two most common situations of architectural design and urban renewal are new design on vacant land or reconstruction based on the original land, in pix2pix, the former is achieved by inputting a blank image into the training source; the latter requires the boundary of the floor plan as the training source. Furthermore, in order to meet the requirements of retaining the grey space hierarchy and functional layout relationship of the generated design, and also to test the influence of the amount of channel information in the image on the pix2pix training process, the training target is divided into greyscale and RGB groups. First, the figure-ground of the sample family is converted into a bubble map, which has 12 rooms and 11 core adjacencies (Figure 1). In the greyscale experiment, 12 rooms were assigned 5 grayscales, and in the colour experiment, each room was assigned a different RGB colour. There are 6 core adjacencies in the grayscale experiment, namely "1-3, 2-3, 2-4, 3-4, 3-5, 4-5", and the corresponding quantities are "1,2,1,4,1,2" respectively. In the RGB experiment, since the RGB colour of each room is different, 11 core adjacencies are recorded. Then, according to the above topological relationship, various floor plans are generated in the written dataset generation module (Figure 4). In addition, two sub-evaluation modules, pre-evaluation and recheck, are added before and after the dataset generation module to ensure the quality of the generated dataset (Figure 4). Among them, the pre-evaluation sub-module is responsible for adjusting the parameters of the generation module to ensure that the input topological structure can be completely retained and read, otherwise, an error will be reported. For example, in this experiment, when generating 1000 datasets, when the component does not report an error and can achieve a high output quantity, the setting of the parameter max adjacency distance should be greater than or equal to 2 (Figure 5_left), and the density parameter should be less than 300 (Figure 5_right). The Recheck module is responsible for detecting that the topological structure in the image is still complete and filtering out unqualified datasets (Figure 6).

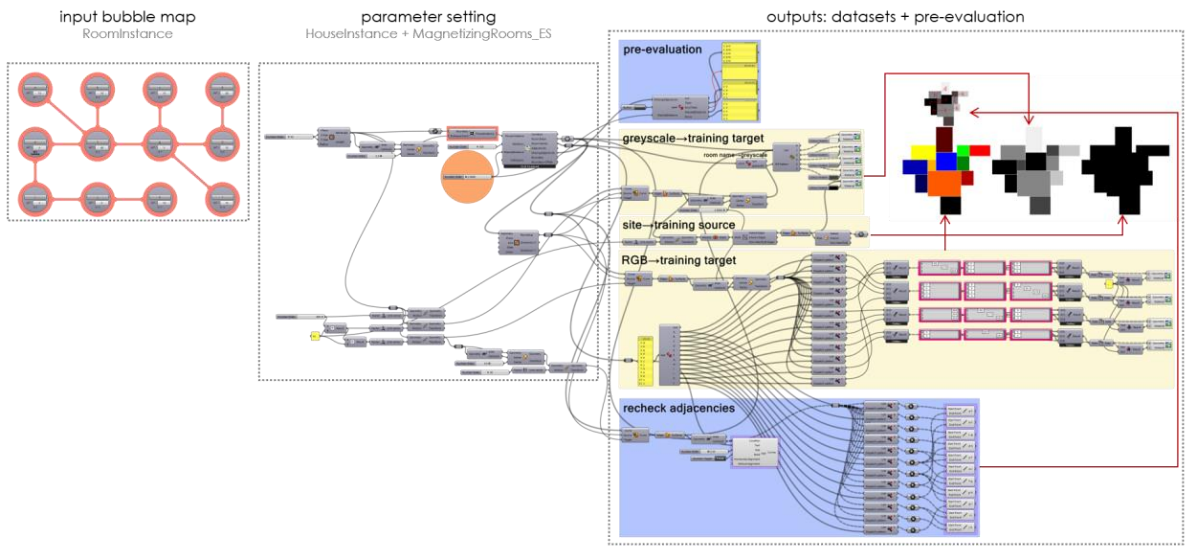

**Figure 4. Generate Datasets in Grasshopper**

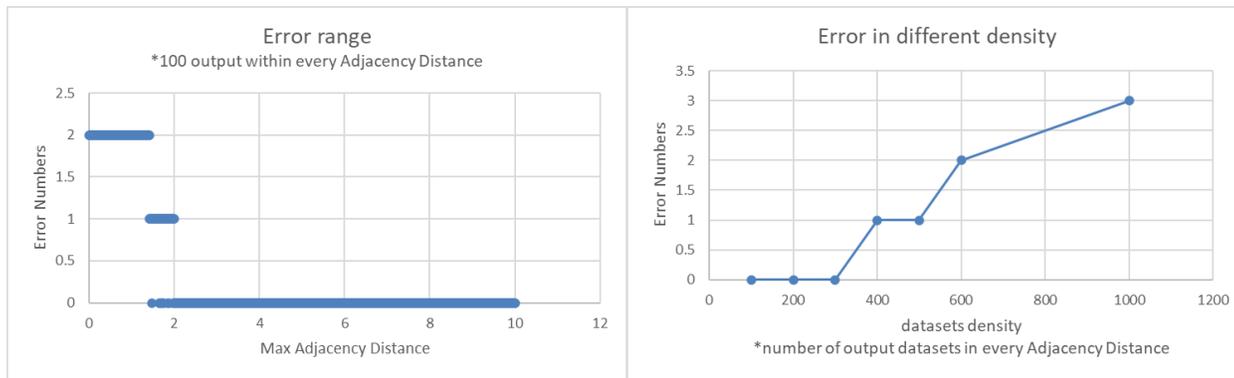

**Figure 5. Pre-evaluation of Datasets**

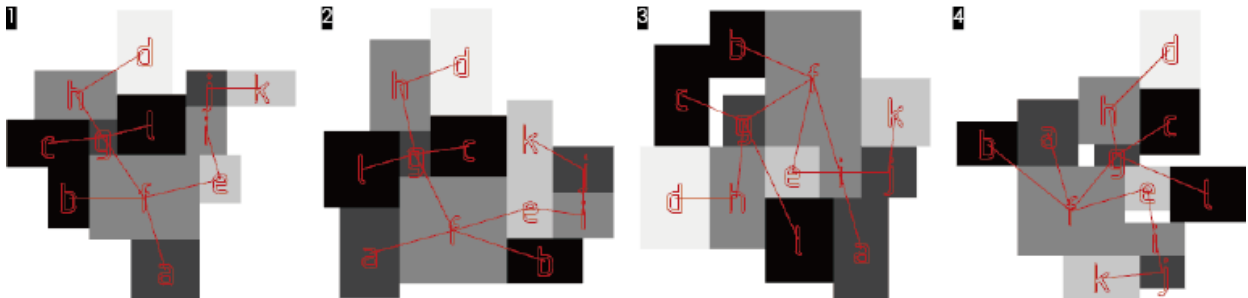

**Figure 6. Recheck Adjacencies of Datasets**

*Note.* Subfigure 1 is a qualified adjacency relationship sample, and subfigures 2 to 4 are unqualified samples. There are three conditions for judging unqualified: 1) Entrance e cannot be surrounded in the middle (subfigure 2); 2) There cannot be gaps or disconnections between adjacent rooms (subfigure 3); 3) The connection relationship between adjacent rooms cannot be just a point (subfigure 4).

### 2.2. GAN Training

The datasets produced in the above steps are input to pix2pix for training. First, each set of datasets is input to the Input Image and Target Image of pix2pix respectively, the learning direction is set from the former to the latter, and the parameters are adjusted according to the pre-training. Second, the storage interval of pix2pix is set. The loss data of each epoch is recorded in the loss_log file; the generated image and training model are output once every certain epoch and recorded in the pth file. These two files will be used for the next loss curve analysis and topology structure learning rate analysis respectively. After the training is completed, the real and fake images compressed in the pth file are separated, and the latter is the training result of the current epoch.

This study is divided into four pieces of training, of which the first two are grayscale trainings and the last two are colour trainings (Figure 1). First, it is hoped to compare the first two groups of experiments to understand the impact of the dataset size on the results. Second, hope to compare the last two groups of experiments to understand the impact of input restrictions on the training process. Third, expect to analyse the impact of the number of image channels on the learning rate of pix2pix by comparing the second and third groups of training. Therefore, except for training1, which has 1000 datasets, the other three experiments have 2500 datasets. The input dataset for the first three trainings is the site boundary, and the fourth is a blank image. The target dataset for the first two groups is rooms with a grey space hierarchy. On the other hand, according to pre-training, it is appropriate to stop training at 500 epochs when the dataset size is 2500. In addition, the minimum number of samples required for topology evaluation is determined. In this experiment, when the number of samples per epoch is at least 50 (2% of the total), almost the same results as evaluating all images can be obtained (Figure 7).

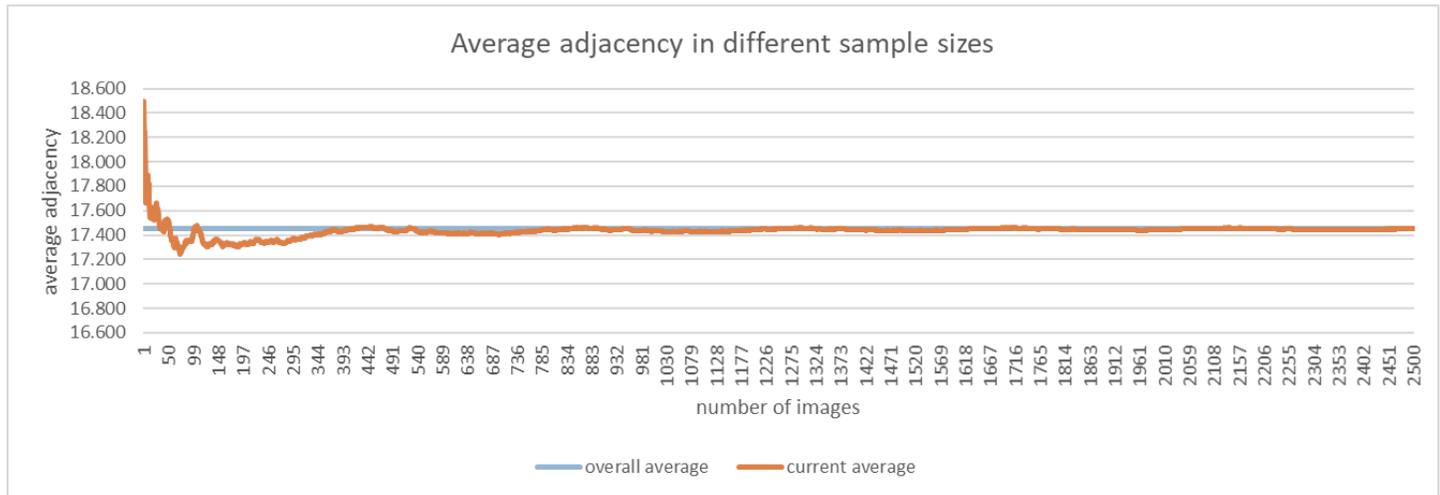

**Figure 7. Minimum Number of Samples Required for Topology Evaluation**

### 2.3. Evaluation

Evaluation is divided into two parts: loss curve and topology evaluation module. On the one hand, by reading the loss_log file, the data (G_GAN, G_L1, D_real, D_fake and total loss) are made into a line graph to intuitively analyze the turning point and stability. The training process of Pix2pix can directly obtain four loss functions, namely G_GAN, G_L1, D_real, and D_fake, and the total loss can be calculated based on them. Among them, G_GAN and G_L1 are generated by the generator part, which allows the generated image to become structurally similar to the target image. In addition, the formula to calculate the total generator loss is GAN loss + LAMBDA x L1 loss, where LAMBDA = 100[23] . And D_real and D_fake are generated by the discriminator part. The discriminator loss function takes 2 inputs: real images and generated images. In simple terms, it is the result of predicting real versus predicting fake. The following introduces the definitions and differences of all five loss functions:

1) G_GAN is the abbreviation of GAN's generator loss, which is a sigmoid cross-entropy loss of the generated images and an array of ones, and can also be simply understood as predict fake.

2) G_L1 refers to L1 loss, which is a MAE (mean absolute error) between the generated image and the target image, which means the result of comparing targets to outputs.

3) D_real refers to real loss, which is a sigmoid cross-entropy loss of the real images and an array of ones (since these are the real images).

4) D_fake represents generated loss, which is a sigmoid cross-entropy loss of the generated images and an array of zeros (since these are fake images).

5) In addition to the above data that can be directly read from the loss_log file output by pix2pix, the total loss can also be calculated, which is the sum of real loss and generated loss[23] .

On the other hand, using the topology evaluation module written in Grasshopper, the spatial adjacency relationships are extracted from the images output by pix2pix, and they are compared and analyzed with the original topological structure of the village sample flat, which is divided into three steps (Figure 8, Figure 9): 1) get brightness or RGB information. Input images' path as a list, reset button and timer, and batch read the brightness or RGB data of each pixel in each image in the list. 2) Cull small areas and scale. Merge pixels in each category into surfaces, then enlarge each room by a certain multiple so that the rooms can intersect. 3) Solid intersection. Count the number of intersections between surfaces and then output all room adjacencies and core adjacencies, and generate a learning rate curve. Finally, analyze when the learning rate curve of pix2pix will reach the original 11 core adjacencies and all adjacencies (here the average value of all adjacencies in the 2500 images in the dataset) and maintain stability.

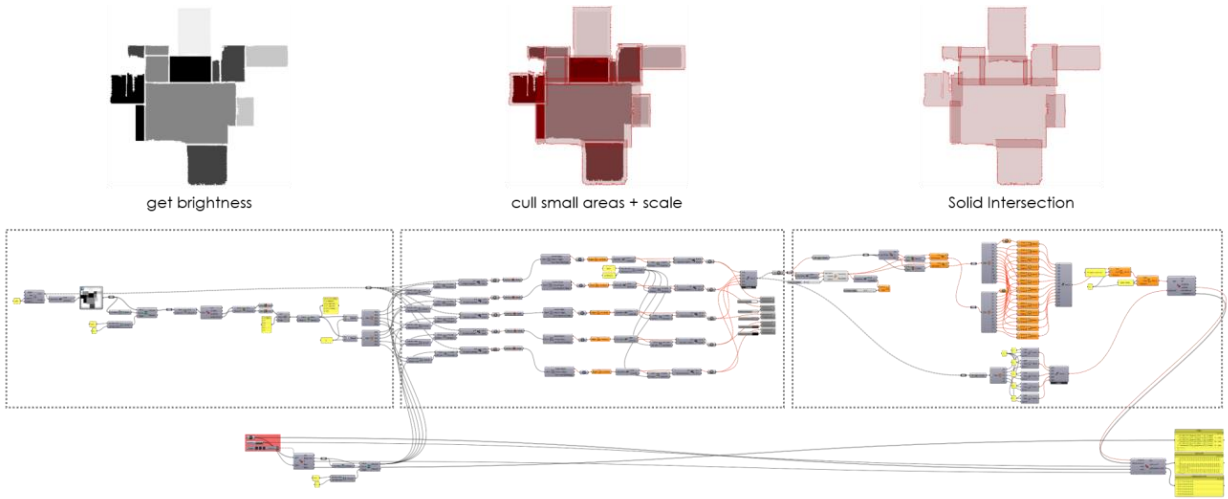

**Figure 8. Greyscale Adjacencies Evaluation Process**

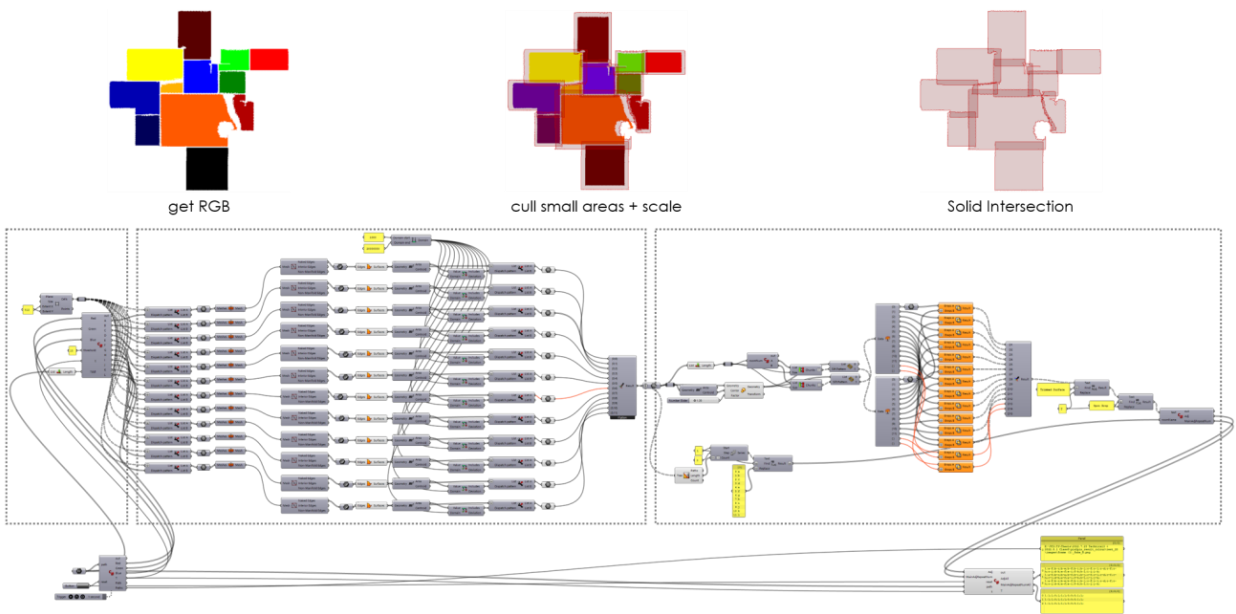

**Figure 9. RGB Adjacencies Evaluation Process**

### 3.  Results & Analysis

#### 3.1.  Results

Figure 10 to Figure 21 show the results of four groups of experiments, mainly observing the following three points: First, observe the changes in the training images. Second, observe when the five loss curves become stable, which is the basis for whether the image is clear and stable. Therefore, for G_GAN and G_L1 from the generator, the main thing is to observe when they turn to stability. For D_real, D_fake and total loss from the discriminator, mainly observe when their noise disappears. Third, observe when the pix2pix learning rate curve reaches the original core adjacencies of the target and its turning point.

Therefore, it was found that the training process of the four groups of experiments can be divided into three periods:

1) Early stage, that is, 0~25epoch of training 1, 0~3epoch of training 2 and 3, and 0~5epoch of training 4. All colour blocks of all images move randomly in all directions within the base range, and the learning rate and G_GAN l oss curve also rise sharply. The core adjacencies of the learning rate reach the original core adjacencies, and even learn all adjacencies in datasets.

2) The middle stage, i.e. 25~100 epochs of training 1, 3~100 epochs of training 2, 3~25 epochs of training 3, and 5~100 epochs of training 4. During this period, the approximate colour positions are determined, the positions of the intermediate colours are gradually reduced and determined, and the edges become clearer from blurry. In addition, the first turning point of G_GAN and G_L1 loss appeared — G_GAN changed from a sharp rise to a slow rise, and G_L1 changed from a sharp drop to basically maintaining stability.

3) The late stage, i.e. after 100 epochs of training 1 and 4, 3~100 epochs of training 2, 3~25 epochs of training 3, and training 4. The amplitude of the image change is very small, but there is still a slight back and forth between the rooms, and the extreme difference of all losses has become smaller and more stable.

Among them, since training 4 has the least restrictions, it is in a continuous and drastic change throughout the process. However, it actually still follows a similar learning process, but the process is slower and more difficult to maintain stability.

### 3.1.1. *Training1: Greyscale Datasets with 1000 Images*

In the calculation method of adjacency evaluation, it is found that pix2pix recognizes the topological structure only based on the colour connection relationship on the surface of the image, unlike human thinking, which infers the function of the room based on the colour. In order to explore the most appropriate way to extract the topological results of the image, the room number and adjacencies of 10 images in each epoch of the total 1000 images were manually counted as a sample test(Figure 12). The name corresponding to each space was inferred based on the grayscale colour, area size, and adjacency relationship of the image, and then the number of rooms and the number of core adjacencies were counted. In addition to the two most obvious disadvantages of low efficiency and great uncertainty, the bigger problem of manual statistics is that even if the uncertain results are counted with the optimal inference, the learning rate of room number and adjacency will remain at a low level from beginning to end. This is contrary to the loss data and the phenomenon that the output image changes from drastic changes to smooth stability. Therefore, the reason is that the wrong statistical method is used, and only the grayscale connection relationship is counted as the method for extracting image adjacencies through programming.

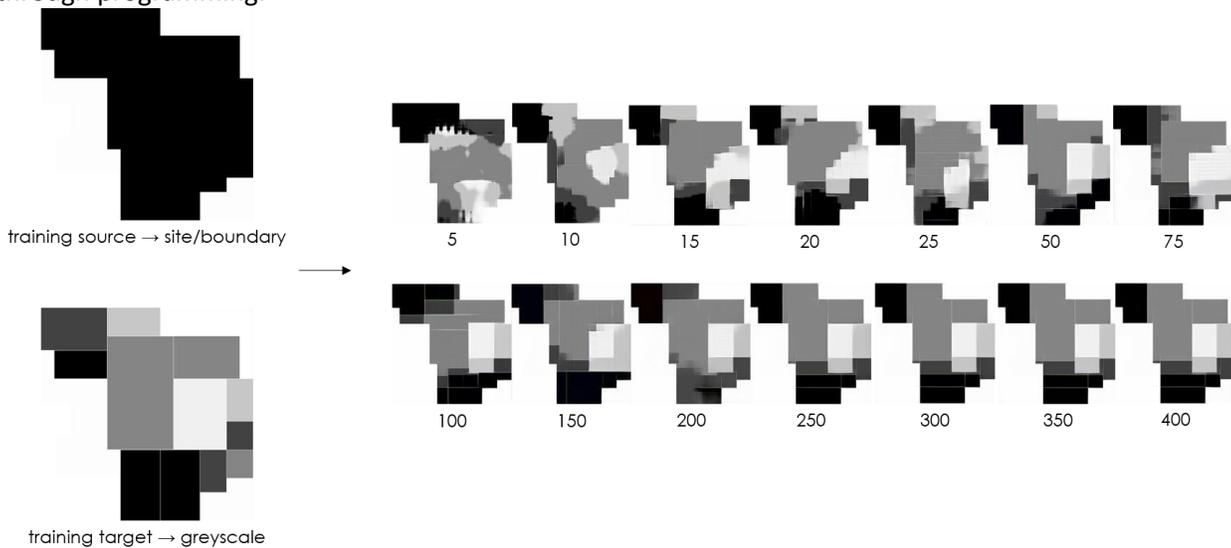

Figure 10. Sample of Training1

*Note*. For seeing more 50 samples, please see Appendix 1

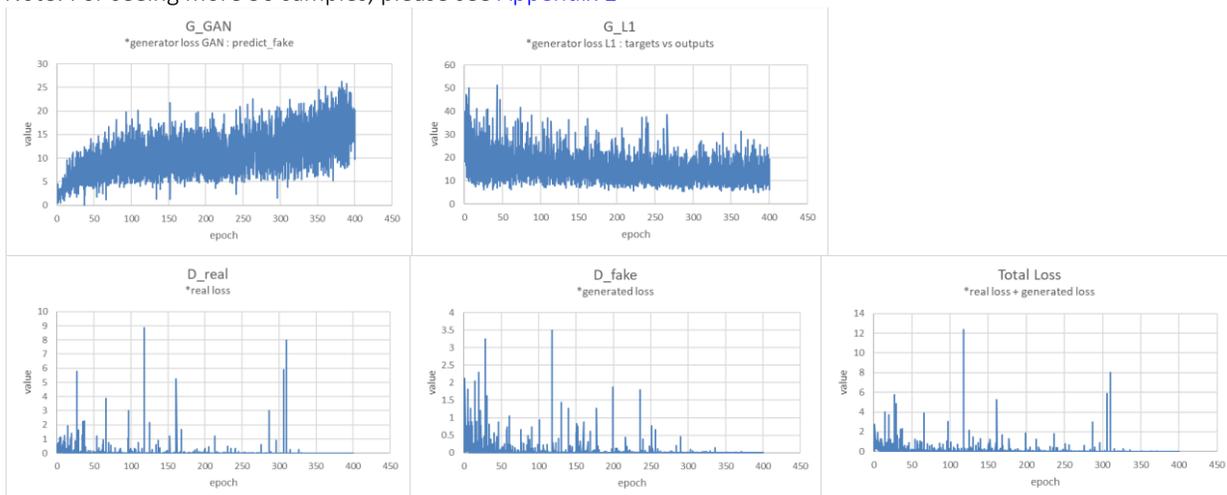

**Figure 11. Loss Curve of Training1**

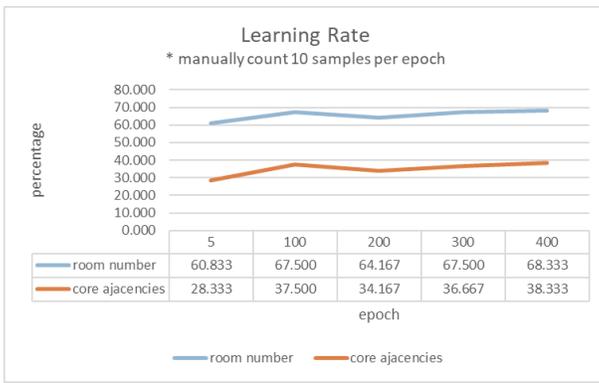

**Learning Rate**
* manually count 10 samples per epoch

| | 5 | 100 | 200 | 300 | 400 |
|---|---|---|---|---|---|
| room number | 60.833 | 67.500 | 64.167 | 67.500 | 68.333 |
| core ajacencies | 28.333 | 37.500 | 34.167 | 36.667 | 38.333 |

epoch

**Figure 12. Manually Calculate Room Numbers and Adjacencies for 10 Images Per Epoch**

### 3.1.2. Training2: Greyscale Datasets with 2500 Images

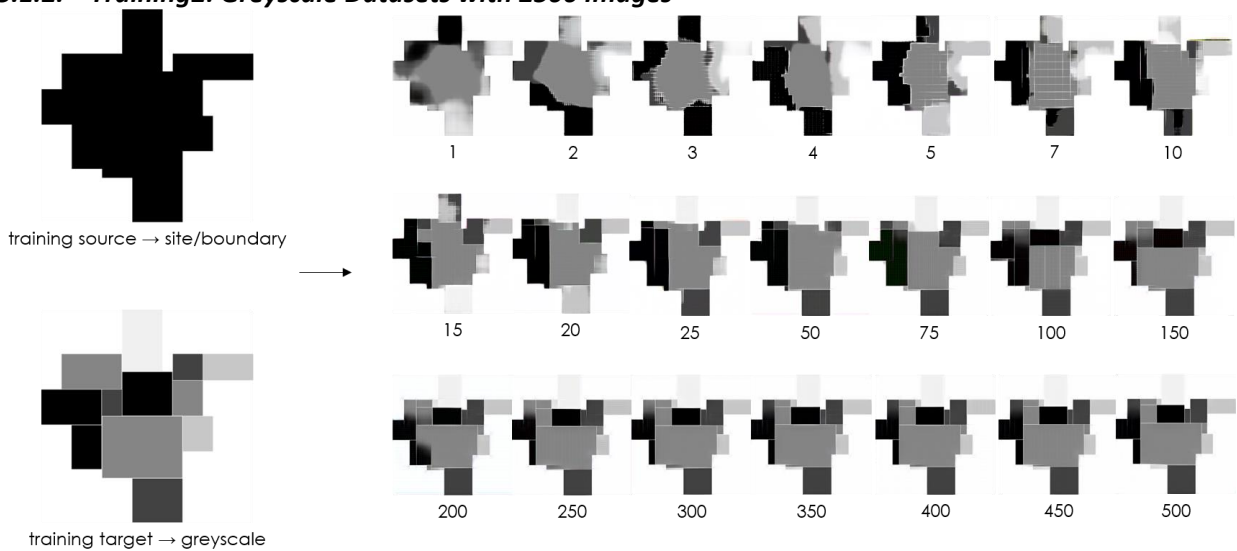

training source → site/boundary

training target → greyscale

**Figure 13. Sample of Training2**

Note. For seeing 50 more samples, please see Appendix 2.

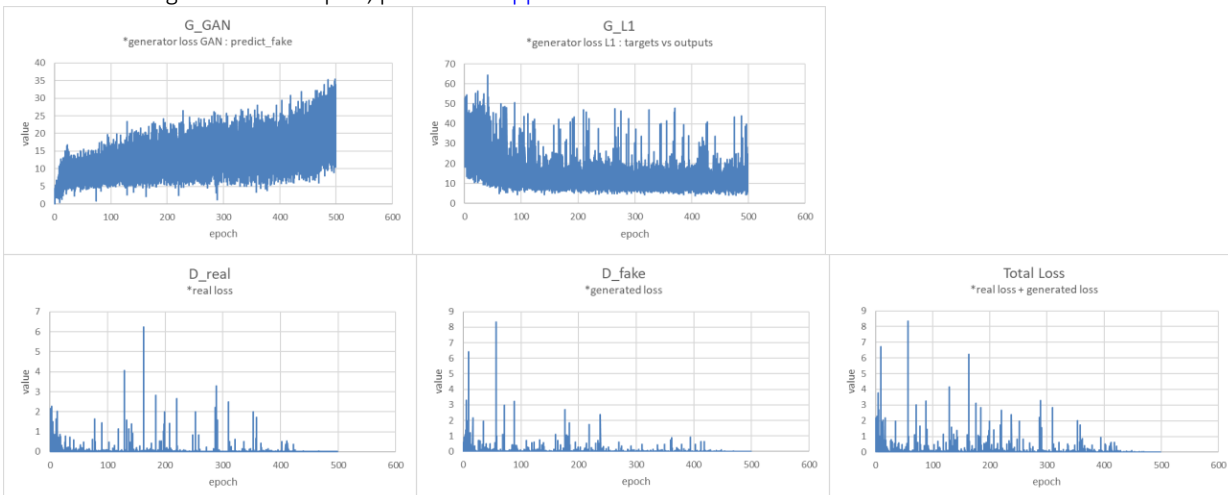

**Figure 14. Loss Curve of Training2**

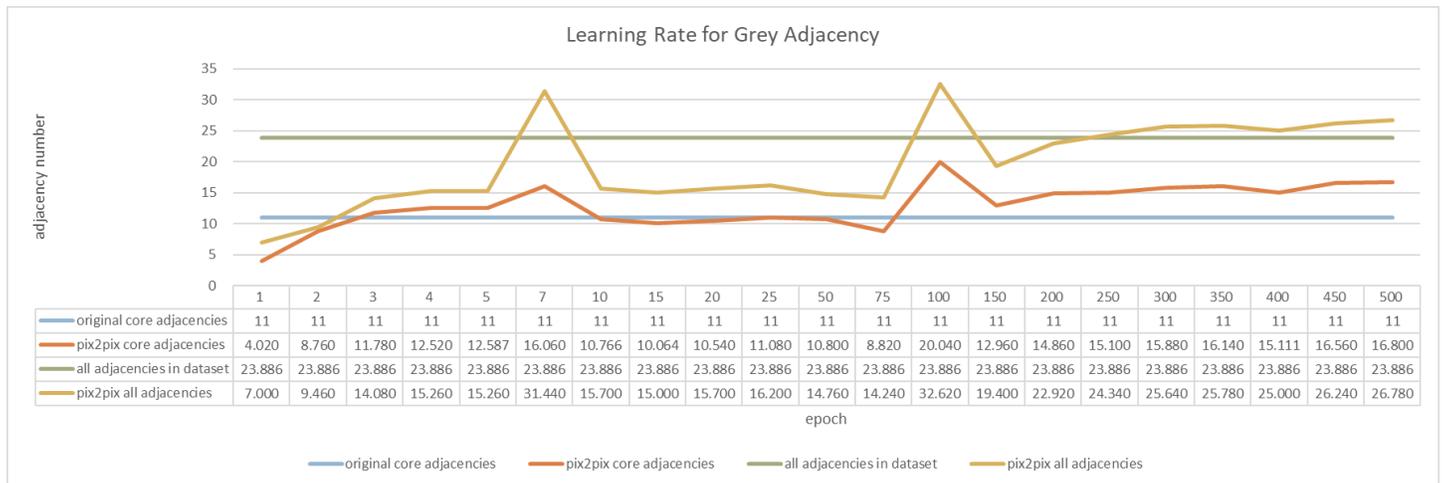

| | 1 | 2 | 3 | 4 | 5 | 7 | 10 | 15 | 20 | 25 | 50 | 75 | 100 | 150 | 200 | 250 | 300 | 350 | 400 | 450 | 500 |
|---|---|---|---|---|---|---|---|---|---|---|---|---|---|---|---|---|---|---|---|---|---|
| original core adjacencies | 11 | 11 | 11 | 11 | 11 | 11 | 11 | 11 | 11 | 11 | 11 | 11 | 11 | 11 | 11 | 11 | 11 | 11 | 11 | 11 | 11 |
| pix2pix core adjacencies | 4.020 | 8.760 | 11.780 | 12.520 | 12.587 | 16.060 | 10.766 | 10.064 | 10.540 | 11.080 | 10.800 | 8.820 | 20.040 | 12.960 | 14.860 | 15.100 | 15.880 | 16.140 | 15.111 | 16.560 | 16.800 |
| all adjacencies in dataset | 23.886 | 23.886 | 23.886 | 23.886 | 23.886 | 23.886 | 23.886 | 23.886 | 23.886 | 23.886 | 23.886 | 23.886 | 23.886 | 23.886 | 23.886 | 23.886 | 23.886 | 23.886 | 23.886 | 23.886 | 23.886 |
| pix2pix all adjacencies | 7.000 | 9.460 | 14.080 | 15.260 | 15.260 | 31.440 | 15.700 | 15.000 | 15.700 | 16.200 | 14.760 | 14.240 | 32.620 | 19.400 | 22.920 | 24.340 | 25.640 | 25.780 | 25.000 | 26.240 | 26.780 |

**Figure 15. Learning Rate for Grey Adjacencies in Training2**

### 3.1.3.  Training3: RGB Datasets with 2500 Images

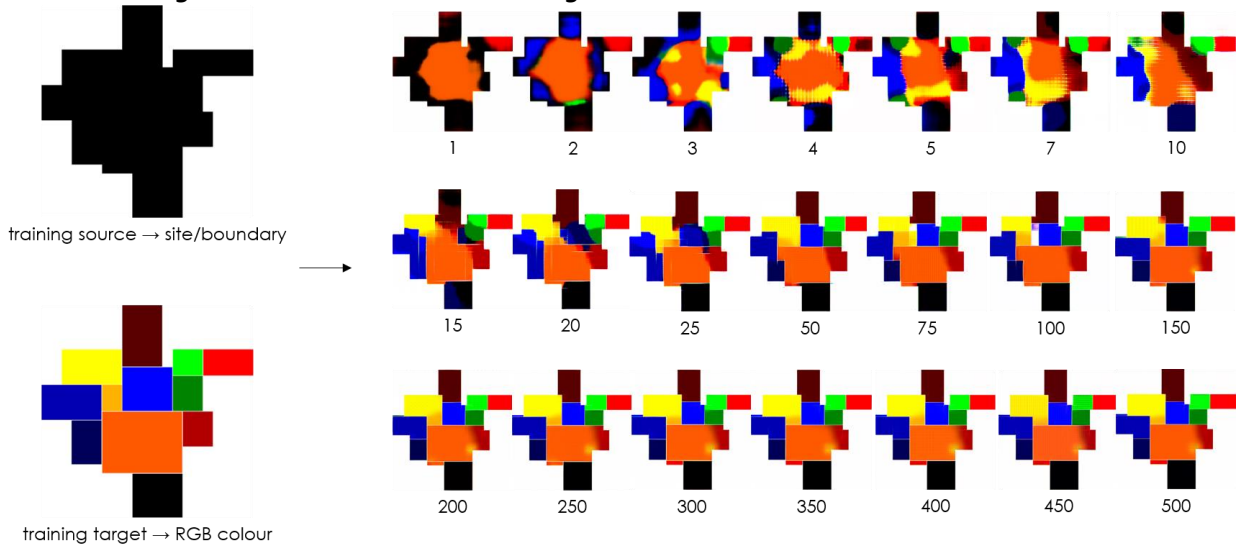

**Figure 16. Sample of Training3**

*Note*. For seeing 50 more samples, please see Appendix 3.

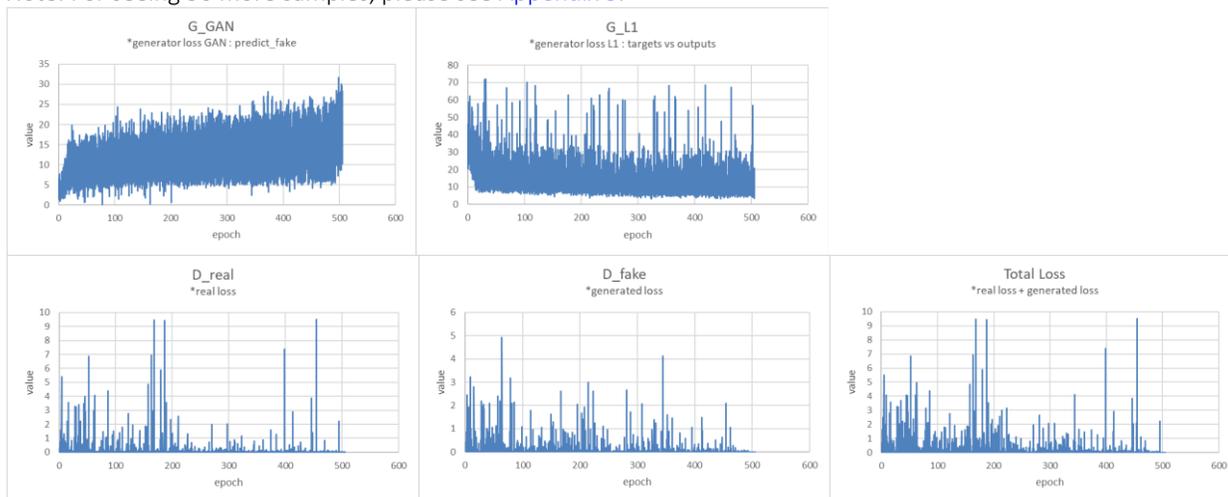

**Figure 17. Loss Curve of Training3**

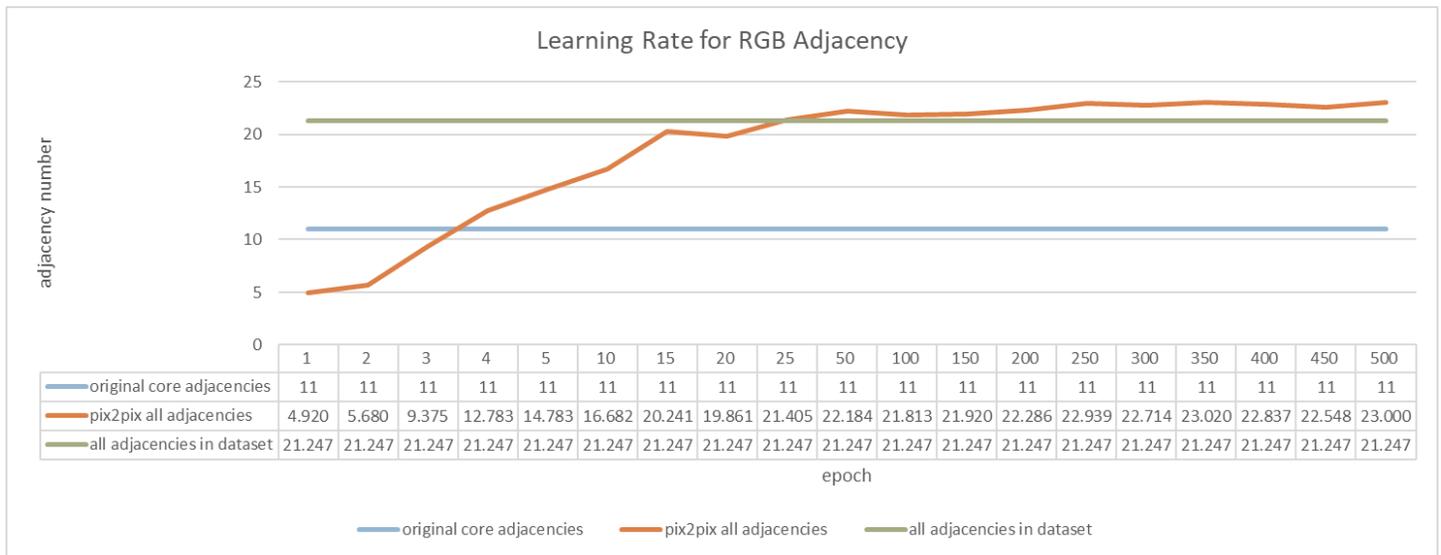

| | 1 | 2 | 3 | 4 | 5 | 10 | 15 | 20 | 25 | 50 | 100 | 150 | 200 | 250 | 300 | 350 | 400 | 450 | 500 |
|---|---|---|---|---|---|---|---|---|---|---|---|---|---|---|---|---|---|---|---|
| original core adjacencies | 11 | 11 | 11 | 11 | 11 | 11 | 11 | 11 | 11 | 11 | 11 | 11 | 11 | 11 | 11 | 11 | 11 | 11 | 11 |
| pix2pix all adjacencies | 4.920 | 5.680 | 9.375 | 12.783 | 14.783 | 16.682 | 20.241 | 19.861 | 21.405 | 22.184 | 21.813 | 21.920 | 22.286 | 22.939 | 22.714 | 23.020 | 22.837 | 22.548 | 23.000 |
| all adjacencies in dataset | 21.247 | 21.247 | 21.247 | 21.247 | 21.247 | 21.247 | 21.247 | 21.247 | 21.247 | 21.247 | 21.247 | 21.247 | 21.247 | 21.247 | 21.247 | 21.247 | 21.247 | 21.247 | 21.247 |

**Figure 18. Learning Rate for Grey Adjacencies in Training3**

### 3.1.4.   Training4: Blank to RGB Datasets with 2500 Images

The fourth training is mainly used as a control experiment, whose purpose is to compare with the former three experiments to observe what kind of training results pix2pix can bring when there is no boundary as a training source restriction. From the output images and loss data of the training, it can be found that the whole training process is in a drastic change. In terms of output images, when comparing different images of the same epoch horizontally, it is found that these 2500 images are almost identical. However, when comparing the changes of the same sample in all epochs vertically, its appearance is constantly changing from beginning to end (Figure 19), unlike the previous training, which will gradually stabilize from drastic changes. However, a careful comparison of the trend of image changes shows that it generally follows the three learning processes of pix2pix summarized before, but because the boundary restriction is lost, the conditions for achieving stability are also lost in the second half. In a nutshell, the loss of the training source will prevent pix2pix from entering a stable state, but even without it, the output image will go through a process from blurring to sharpening both the outer and inner edges.

Secondly, the continuous instability is also reflected in the training loss (Figure 20). The five loss curves of this training are very different from all previous training, especially the two loss curves of G_GAN and G_L1 from the generator data. In all previous pieces of training with site boundaries, the curve of G_GAN is an "S" shape that first rises, then stabilizes, and finally rises or falls. And the range of each epoch shows a trend of gradually increasing (Figure 11, Figure 14, Figure 17). However, their overall upward trend in this training is very gentle, and there are several large fluctuations in the middle. At the same time, the range within the same epoch is maintained in a very narrow range from beginning to end (Figure 20_1). In addition, G_L1 is very different from the previous training curve, but its changes are not the same as G_GAN loss. In previous training, G_L1 showed an "L" shape that was opposite to G_GAN, that is, it dropped sharply in the early stage and stabilized in the later stage, and the extreme changes within the epoch remained consistent throughout the training process. However, in training 4, G_L1 lost all curve changes and remained between 60 and 100 values, which was much higher than all previous training, like a very wide straight line (Figure 20_2). On the other hand, the performance of D_real and D_fake in training 4 was similar to that of the previous experiments, and the fluctuation of value basically disappeared after about 400 epochs (Figure 20_3,4).

Therefore, the similarities and differences of these five loss data further illustrate the connection between each of them and the result images. The two data from the generator of G_GAN and G_L1 are mainly closely related to whether the image is stable. If pix2pix is unable to produce stable results after a long period of operation like this experiment due to the lack of certain restrictions, then G_GAN will not be able to show an "S" shape but a fluctuating shape, and the G_L1 curve will not show an "L" shape. On the other hand, the three items of data from the discriminator, D_real, D_fake, and total, are mainly related to the clarity of the image. As long as the images are clear without noise or blur, their range will decrease and eventually disappear. Therefore, this experiment shows that based on the changes in the loss curve alone, it is possible to first determine what problems may have occurred in the training process before exporting the training results.

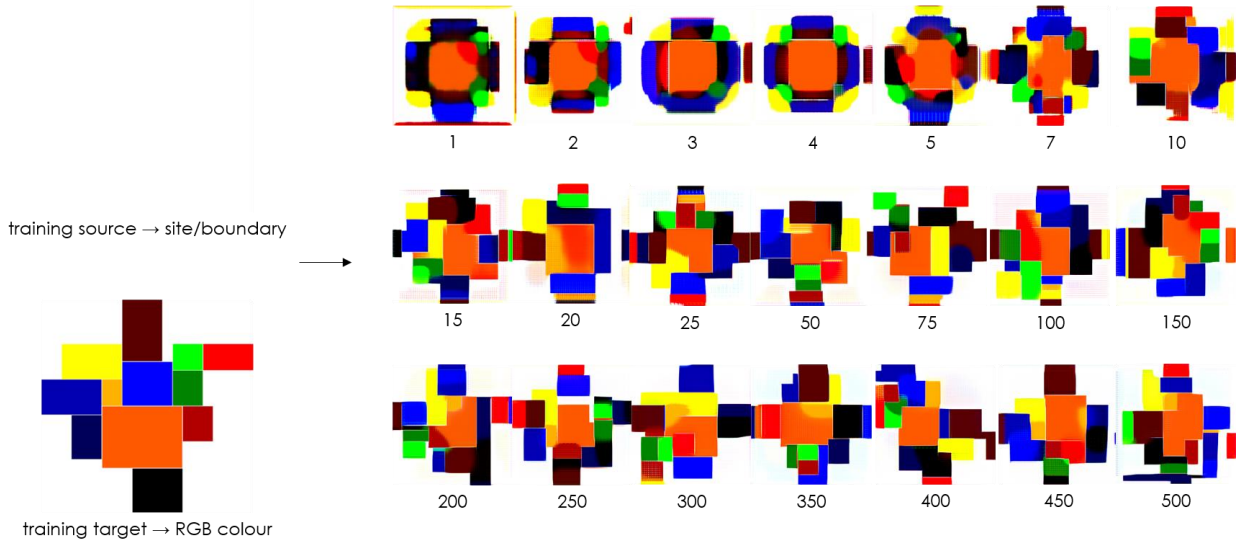

training source → site/boundary

training target → RGB colour

**Figure 19. Sample of Training4**

Note. For seeing 50 more samples, please see Appendix 4.

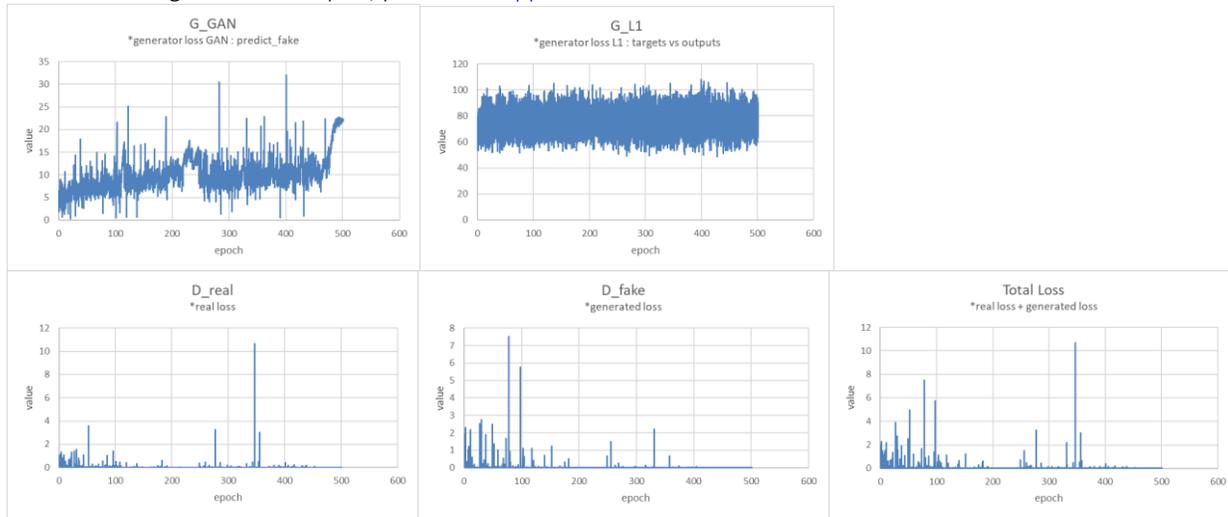

**Figure 20. Loss Curve of Training4**

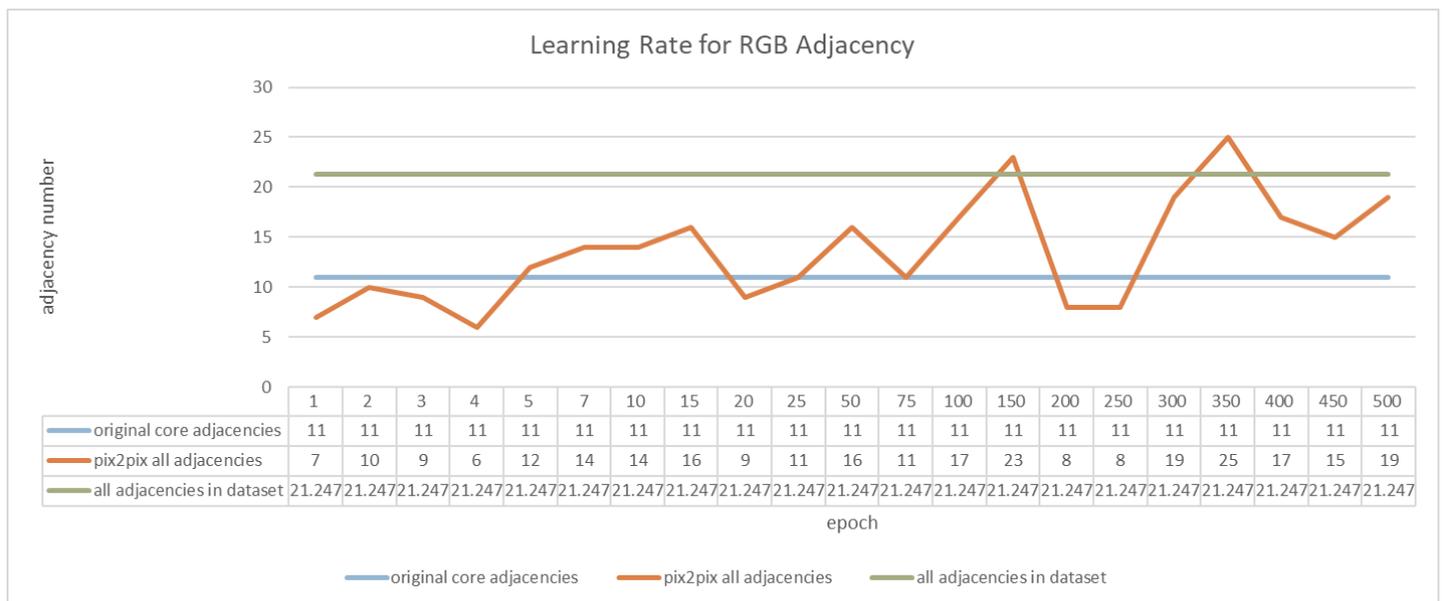

**Figure 21. Learning Rate for Grey Adjacencies in Training4**

### 3.2.    Analysis

#### 3.2.1.    Training1 vs 2: Datasets volume

By comparing the first and second training, we can know the impact of the amount of dataset on the training results. It is found that the amount of dataset mainly affects the curve change and training time. When the dataset is enlarged, the turning point and curve change of G_GAN and G_L1 loss curves are more obvious. And D_real and D_fake need more epochs to reach stability, but when the training reaches the later stage of the stable stage, the change is very small (Figure 11, Figure 14). In addition, in the calculation method of adjacency evaluation, it was found that the pix2pix cognitive topology structure was only based on the colour connection relationship on the image surface, unlike human thinking which would infer the function of the room based on colour.

#### 3.2.2.    Training3 vs 4: Training Source

The comparison between the third and fourth training can help us understand the impact of the training source on the results. First, losing the training source will prevent pix2pix from entering a stable state and make all outputs homogenized, but even so, it can still learn slowly. Therefore, the role of the training source of pix2pix is a topological definition and range division. Second, the continuous instability is also reflected in the training loss, especially from the generator G_GAN and G_L1 (Figure 21). In the previous three trainings, G_GAN showed an "S" shape that first rose, then stabilized, and finally rose or fell, and the range of each epoch showed a trend of gradually increasing (Figure 11, Figure 14, Figure 17). However, it rose gently in training 4 and was interspersed with several large fluctuations with the range in a very narrow range (Figure 20_1). In addition, G_L1 showed an "L" shape in the previous training — a sharp drop in the early stage and stability in the later stage. However, in training 4, G_L1 lost all its curve changes and maintained a wide straight line between 60 and 100 values, which is much higher than all previous training (Figure 20_2). The performance of D_real and D_fake in training 4 is close to that of the previous ones, and the fluctuation of value basically disappears at about 400 epochs (Figure 20_3,4). Therefore, the similarities and differences of these five loss data further illustrate the relationship between each of them and the result images. On the one hand, G_GAN and G_L1 from the generator are mainly closely related to whether the image is stable. If the lack of constraints leads to the inability to determine a stable result after a long period of operation, then G_GAN will not appear "S" type but fluctuate, and the G_L1 curve will not show "L" type. On the other hand, D_real, D_fake and total, the three data from the discriminator, are mainly related to the clarity of the image. As long as there is no noise or blur, their range will decrease and finally disappear. Therefore, based solely on the changes in the loss curve, we can first identify problems related to image quality before exporting training results, but we cannot predict the learning rate at the topological level.

#### 3.2.3.    Training2 vs 3: Greyscale vs RGB

The purpose of the comparison between training 2 and 3 is to compare the differences in training results when the training target is grayscale and colour. It is found that the learning process of grayscale and RGB experiments is actually very similar, the only difference is the order of the turning point. From this, we can get two characteristics of pix2pix in topological learning:

On the one hand, the turning point of the loss curve of the RGB experiment is more obvious, indicating that more data will make the computer have a clearer training change. Overall, the adjacency learning rate of both groups presents an "S" shape and can be divided into three similar stages. They all learned the original core adjacencies in about 3 epochs, but the learning of the RGB group is more continuous and closer to the shape of an inverted "L" (Figure 18). The learning speed of the grayscale experiment is slower and longer, and it will stagnate in the middle stage of the tug-of-war between small colour blocks. Therefore, in the middle stage, its two sub-stages can be further observed: first, determine the position of the fuzzy area, and then clarify the edge (Figure 16).

On the other hand, the adjacency learning rate in the RGB group rises faster, but the D_real and D_fake losses stabilize more slowly than in the grayscale group. However, in general, it is expected that colour training should get better results than grayscale training because the computer has more information. Therefore, there are two reasons for this inconsistency in the analysis:

First, the adjacency learning rate is the recognition of topological relationships, while the loss of pix2pix, as an image-based generation, only indicates whether the image is clear and whether there are noise points. Therefore, when learning topological structures, although the loss can reflect certain output information, it cannot be used as a complete basis for judgment. In addition, the machine has more data in the RGB dataset than the grayscale, so the adjacency learning rate will rise faster. Therefore, combining the adjacencies learning rate line chart and the actual general results of output images, it is judged that RGB input can indeed make pix2pix achieve higher learning efficiency than grayscale.

Second, the uncertain areas of grayscale images may be misidentified by pix2pix, so the processing of noise and clarity will take more time. The discriminator in the cGAN used by pix2pix is a convolutional PatchGAN classifier that attempts to classify whether each image patch is real or not. Therefore, in the grayscale experiment, since the colour types in the grayscale datasets are fewer and single colours, the blurred area between every two grayscales with large differences may be recognized as a room with an intermediate grayscale in the next epoch. Therefore, the single colour may inadvertently speed up the computer's area recognition and classification, allowing noise to be eliminated faster or unrecognizable, resulting in faster stabilization of D_real and D_fake loss. However, for the RGB experiment, the mixed colour between each two rooms cannot be similar to the colour of other rooms, so the blurred area and noise will be accurately identified and will not be recognized and misclassified by the computer, which will be clearly reflected in the loss.

## 4. Discussion

In terms of technology, this study provides quantitative data support for the topological structure learning ability of pix2pix, and confirms that pix2pix has the potential to pursue high simulation and topological legacy at the same time. Therefore, it can guide the development of GAN to a certain extent in the direction of wider versatility and the integration of multiple image generation needs. In the latest research, the application of machine learning in the field of architecture is still technology-oriented and purposeful. For example, with the improvement of computing performance, GNN has been able to move towards larger urban scales in applications, such as land use planning[44] , road planning[46], urban design[40] and transportation[28] etc. However, how to solve the problem of retaining regional characteristics and integrating 'extrinsic' and 'intrinsic' properties of space in machine learning has been ignored. In addition, the design generation direction is still mainly based on new designs from scratch, with few optimizations and updates based on existing local contexts. Therefore, this study also lays the foundation for the possibility of GAN integrating multiple functions in the future in terms of data and analysis.

In terms of application, this study can promote the universality, regional characteristics and scalability of I2I GAN in architectural applications. First, by simplifying tools and enhancing stability for architectural application problems, this paper will help promote the popularization of GAN in architectural applications. In the past, image-based GAN was limited by the data set in the hands of architects and was mostly used for the stylization of pictures, but could not really meet the needs of design in spatial operations. The data set generation module proposed in this study can extract the spatial topological relationship in the floor plan and convert it into a functional layout or grey space layout, and can also customize the topological relationship to generate a customized image data set. The simplification of complexity can not only save time and help build a bridge of communication between architects and machines but also popularize it to the public and users to achieve co-design. Second, this paper confirms the feasibility of using image-based GAN to preserve spatial topological structures, which can then be used to preserve regional spatial characteristics. Third, it can be applied to a variety of design problems and has scalability. In the experiment, pix2pix can learn topological structure with or without site boundary input restrictions, which can be used for architectural design or urban renewal with or without site restrictions. In addition, the scalability of the application scale will depend on the complexity of the dataset, which can be generated through the dataset generation module of this study.

## 5. Conclusions

In order to assist GAN in generating regional features that take into account both 'intrinsic' and 'extrinsic' properties of space, this paper proposes a method to quickly detect the ability of pix2pix to learn topological relationships by adding generation and detection modules before and after GAN. The results not only confirm the autonomous learning ability of pix2pix, a representative of Image-to-image Translation, but also its learning speed and key information retention rate. Three sub-questions were answered:

First, can GANs, especially pix2pix, autonomously learn the topological structure of images without topological cues? pix2pix can learn the entire topological structure of an image through the colour-adjacent relationship of the image without topological cues. The process can be divided into three stages (Figure 22):

1) Drastic changes in the field: Determine the area with the largest area and the most obvious features, and the completion of topological learning of origin important adjacencies will correspond to the end of this period.

2) Small-scale domain changes and clear edge period: When the image information decreases, the training will mainly stagnate at this stage, and it is divided into two sub-steps: determining the fuzzy area of the image and clearing the edge. Their end corresponds to the turning point of G_GAN and G_L1 respectively. At this stage, the topological learning rate finally rises to all adjacencies and enters a stable turning point.

3) Stable period: At this time, all output images, losses and learning rates are basically stable. D_real, D_fake, and total loss, which indicate whether there are noise points in the image, are also greatly reduced.

Second, how effective is it in learning topological structures? Pix2pix is still very efficient in learning topological relationships. With the number of datasets of 2500 in this study, pix2pix can learn core adjacencies within 3~5 epochs on average, and learn all adjacencies within 25~100 epochs.

Third, what is the impact of greyscale and RGB modes on its learning process? Pix2pix is more efficient in learning topological structures in colour training than in the grayscale group, that is, the more information contained in datasets, the faster it learns. Moreover, if the dataset setting lacks some restrictions (such as the site boundary as the input image datasets), pix2pix can also learn, but the learning efficiency is lower at this time, and stable images cannot be obtained. The learned content is constantly lost and slowly growing.

This paper has two innovations: 1) It proves that pix2pix can autonomously learn spatial topological relationships and apply them to architectural design. 2) It fills the gap in the performance of Image-based Generation GAN from a topological perspective. In addition, the detection method proposed in this study takes a short time and is simple to operate. The two detection modules can be widely used for customizing image datasets with the same topological structure and batch detecting the topological relationship of images. In the future, this paper may provide a theoretical foundation and data support for the application of architectural design and urban renewal that use GAN to retain spatial topological characteristics.

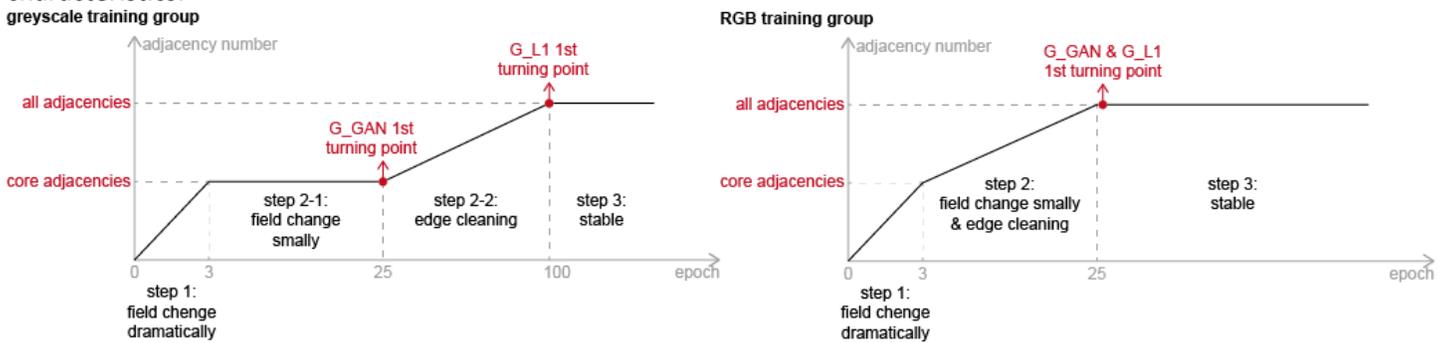

**Figure 22. Conclusion Diagram**

# 6.    Author Contributions

Conceptualization, Yayan Qiu; Data curation, Yayan Qiu; Formal analysis, Yayan Qiu; Investigation, Yayan Qiu; Methodology, Yayan Qiu; Resources, Yayan Qiu; Software, Yayan Qiu; Supervision, Sean Hanna; Validation, Yayan Qiu; Visualization, Yayan Qiu; Writing – original draft, Yayan Qiu; Writing – review & editing, Yayan Qiu and Sean Hanna. All authors have read and agreed to the published version of the manuscript.

**Appendix 1. 50 Samples of Training1**

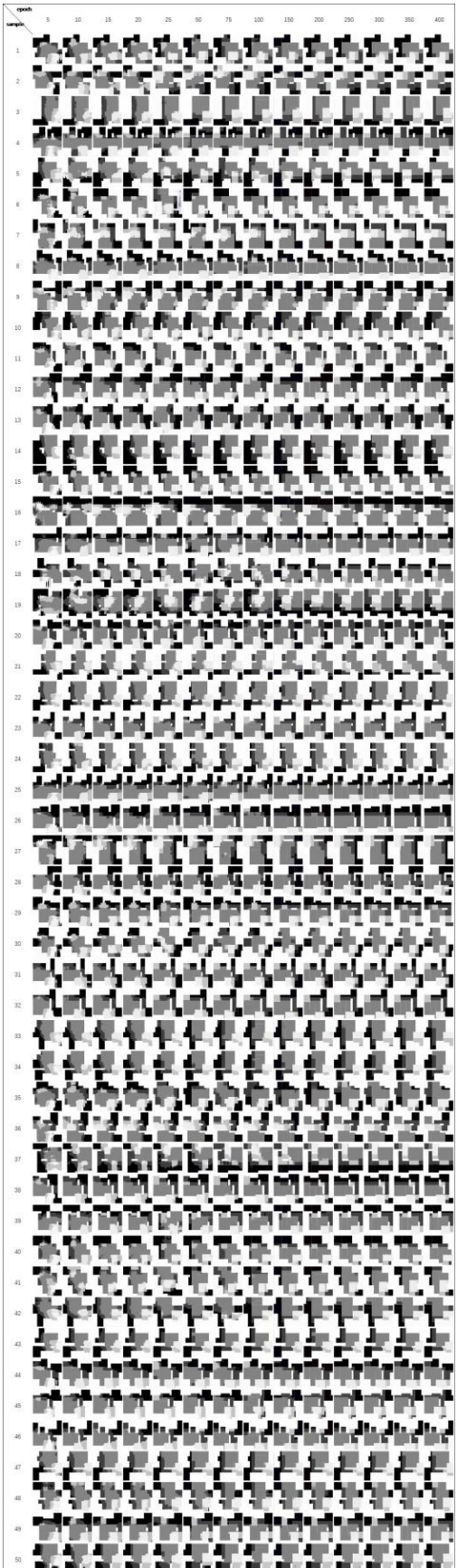

Note. Variation of 50 samples of training 1 over 500 epochs.

**Appendix 2. 50 Samples of Training2**

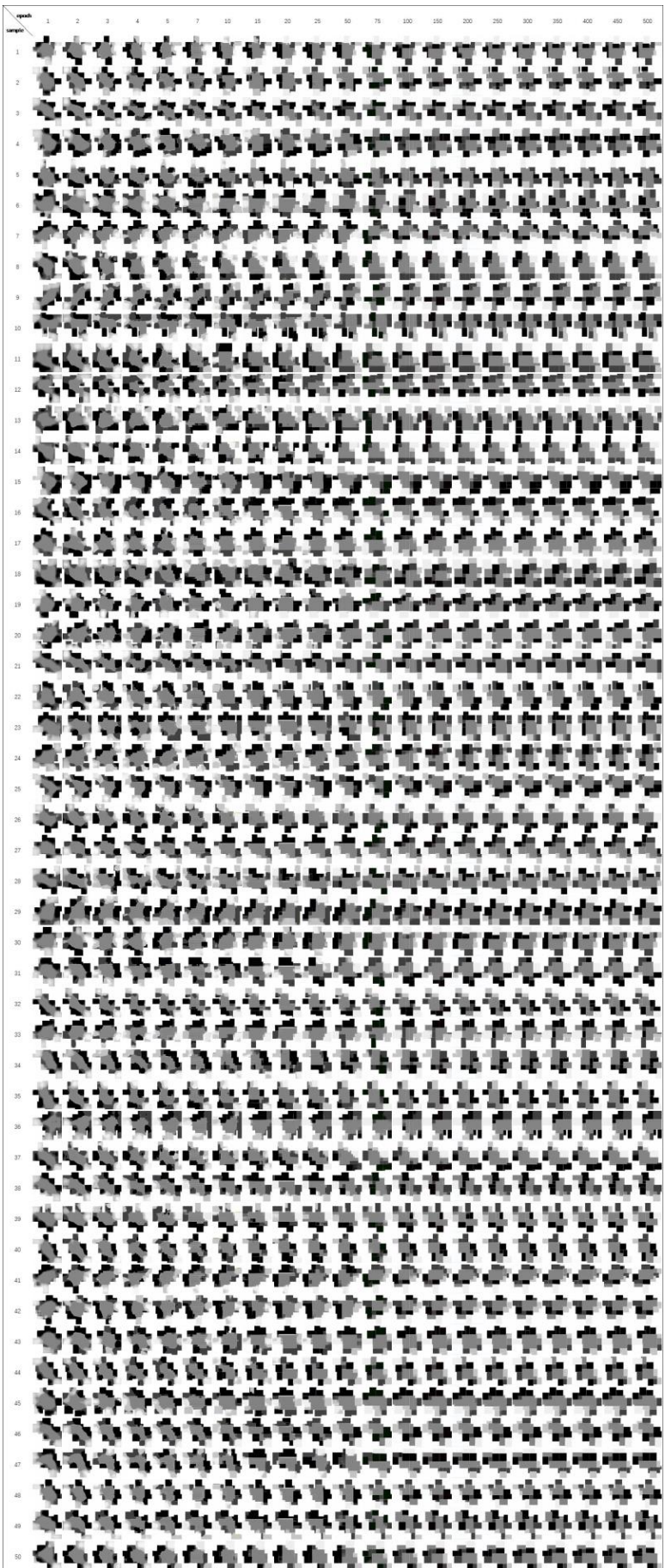

Note. Variation of 50 samples of training 2 over 500 epochs. They are also the 50 samples used in the adjacency evaluation.

**Appendix 3. 50 Samples of Training3**

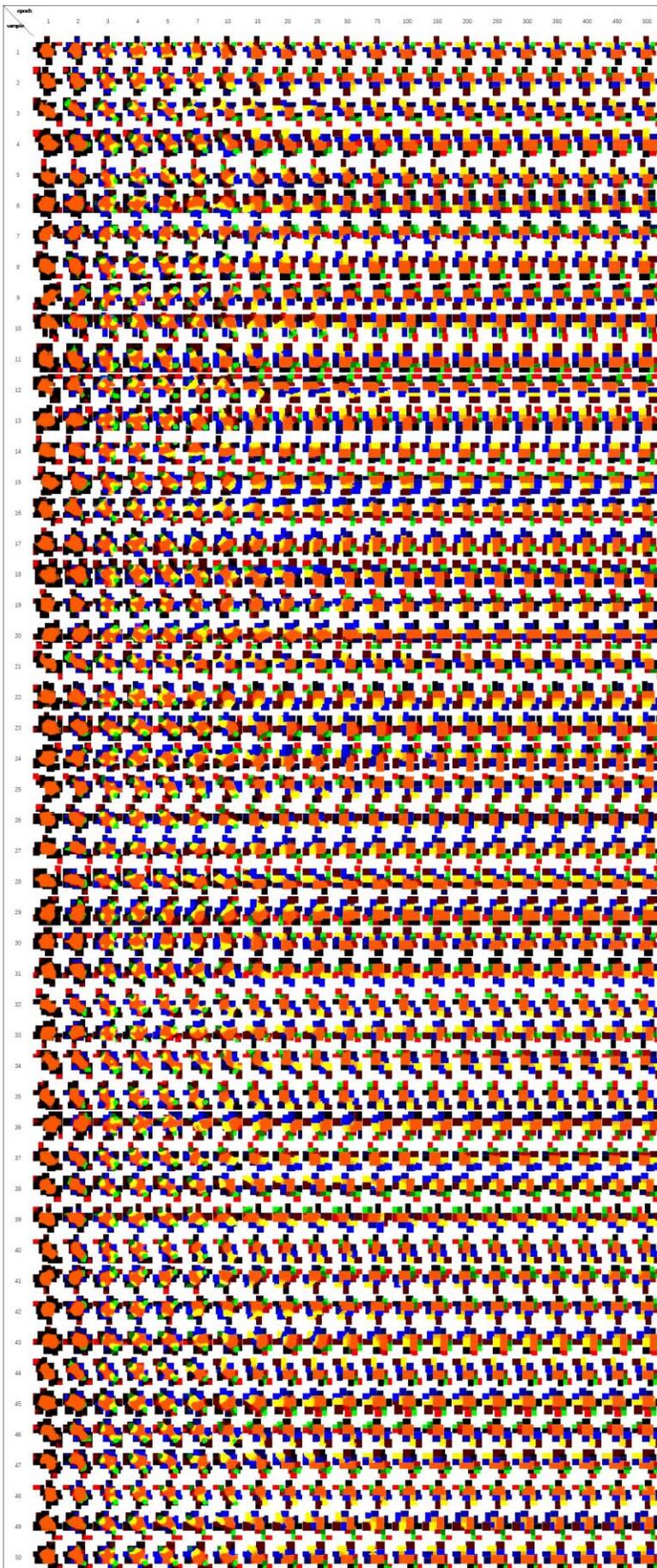

Note. Variation of 50 samples of training 3 over 500 epochs. They are also the 50 samples used in the adjacency evaluation.

**Appendix 4. 50 Samples of Training4**

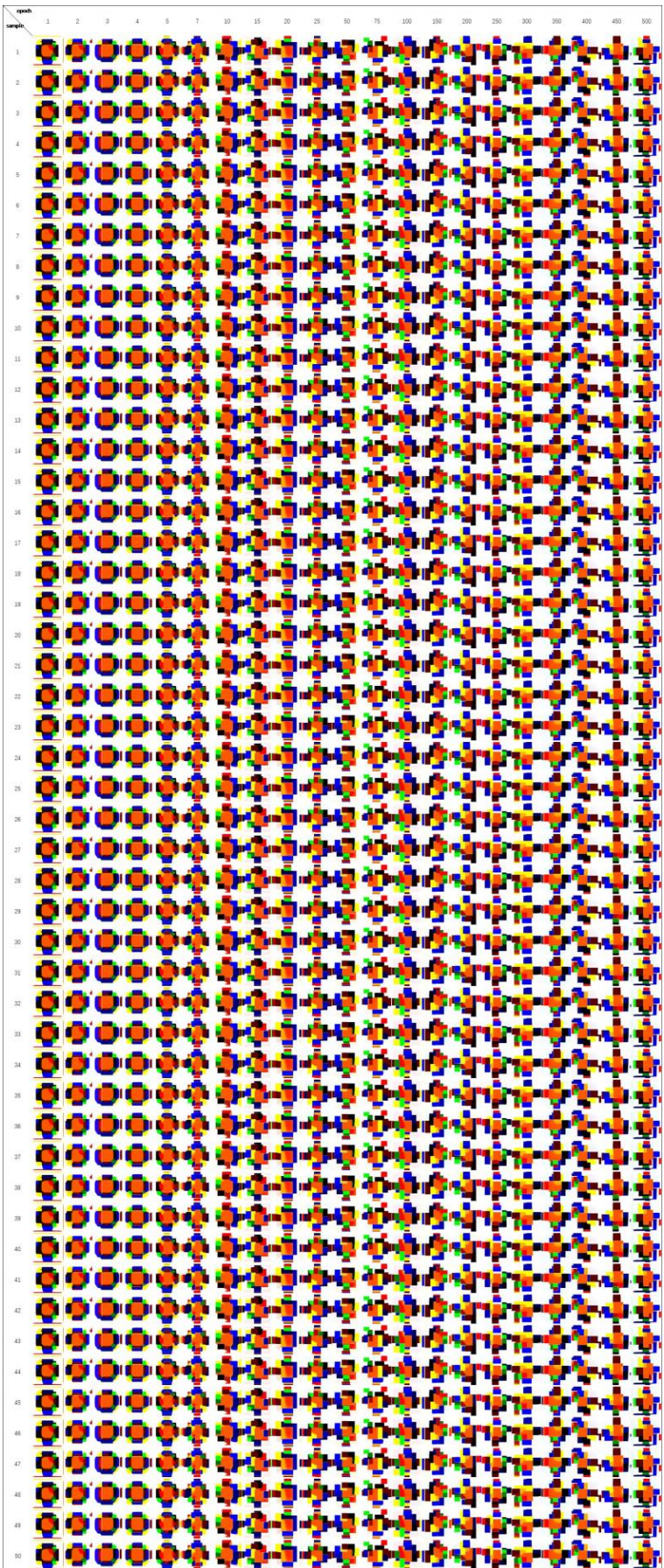

Note. Variation of 50 samples of training 4 over 500 epochs. They are also the 50 samples used in the adjacency evaluation. Since there is no site boundary as the training source, all the samples in each epoch here are the same.